\documentclass{article}

\usepackage[preprint]{neurips_2020}

\usepackage[utf8]{inputenc} % allow utf-8 input
\usepackage[T1]{fontenc}    % use 8-bit T1 fonts
\usepackage{hyperref}       % hyperlinks
\usepackage{url}            % simple URL typesetting
\usepackage{booktabs}       % professional-quality tables
\usepackage{amsfonts}       % blackboard math symbols
\usepackage{nicefrac}       % compact symbols for 1/2, etc.
\usepackage{microtype}      % microtypography

\usepackage{graphicx}
\usepackage{amssymb}
\usepackage{latexsym}

\usepackage{amsmath}
\usepackage{bm}
\usepackage{threeparttable}
\usepackage{algorithm}
\usepackage{algpseudocode}

\title{Characteristics of networks generated by kernel growing neural gas}

\author{
Kazuhisa Fujita\\
Komatsu University, Komatsu, Ishikawa, Japan\\
\texttt{kazu@spikingneuron.net}
}

\date{}

\begin{document}

\maketitle

\begin{abstract}
    This research aims to develop kernel GNG, a kernelized version of the growing neural gas (GNG) algorithm, and to investigate the features of the networks generated by the kernel GNG. The GNG is an unsupervised artificial neural network that can transform a dataset into an undirected graph, thereby extracting the features of the dataset as a graph. The GNG is widely used in vector quantization, clustering, and 3D graphics. Kernel methods are often used to map a dataset to feature space, with support vector machines being the most prominent application. This paper introduces the kernel GNG approach and explores the characteristics of the networks generated by kernel GNG. Five kernels, including Gaussian, Laplacian, Cauchy, inverse multiquadric, and log kernels, are used in this study. The results of this study show that the average degree and the average clustering coefficient decrease as the kernel parameter increases for Gaussian, Laplacian, Cauchy, and IMQ kernels. If we avoid more edges and a higher clustering coefficient (or more triangles), the kernel GNG with a larger value of the parameter will be more appropriate.
\end{abstract}

%\key{Growing neural gas, kernel method, self-organizing map}

\section{Introduction}

Today, the amount of data has grown enormously \cite{Cabanes:2012,Rossi:2014}. To efficiently process such massive datasets, vector quantization methods are often used to reduce the number of data points. Thus, vector quantization methods have become increasingly important for handling large datasets.

Self-organizing maps (SOMs) and their alternatives are widely used vector quantization methods commonly applied in various fields such as data visualization, feature extraction, and data classification. These approaches encode a dataset into a set of interconnected units. Kohonen's SOM is the most popular and widely used of these methods. Kohonen's SOM creates a network with fixed topology, such as a $d$-dimensional lattice.

The growing neural gas (GNG) proposed by Fritzke \cite{Fritzke:1995} is an alternative to SOM. GNG can flexibly change the network topology during training. GNG can adapt not only the reference vectors but also the network topology to an input data set. GNG can gradually increase the number of neurons and reconstruct the network topology according to the input data. GNG is a useful method for extracting the topology of the input data. Thus, GNG can not only quantize a dataset but also preserve the topology of the dataset as the topology of the network.

The kernel method is useful for projecting data into a high-dimensional feature space. The support vector machine \cite{Cortes:1995} gains good performance for non-linear data, applied kernel method. The kernel Kohonen's SOM can perform better than Kohonen's network \cite{Chen:2009a}. However, a kernel version of GNG has not yet been developed, and the characteristics of networks generated by kernel GNG remain unknown.

This study aims to develop the kernel GNG and investigate the characteristics of networks generated by the kernel GNGs with Gaussian, Laplacian, Cauchy, inverse multiquadric (IMQ), and log kernels. First, the method of the kernel GNG is derived as shown in Sec.\ref{sec:kernel_gng}. Second, the paper shows the feature of networks generated by the kernel GNGs with these kernels in Sec.\ref{sec:results}. This paper shows that the kernel GNGs with these kernels can generate networks that effectively represent input datasets, similar to the original GNG algorithm.

\section{Related work} % (fold)
\label{sec:related_work}

The best-known and most widely used the SOM is Kohonen's SOM. The SOM can project multidimensional data onto a low-dimensional map \cite{Vesanto:2000}. The SOM is used in various applications such as color quantization \cite{Chang:2005,Rasti:2011}, data visualization \cite{Heskes:2001}, and skeletonization \cite{Singh:2000}. The most popular SOM is Kohonen's SOM \cite{Kohonen:1982}. However, the network structure generated by Kohonen's SOM is static (generally an $d$-dimensional lattice) \cite{Sun:2017}. Thus, Kohonen's SOM cannot flexibly change the network topology depending on the input dataset.

The growing neural gas (GNG) \cite{Fritzke:1995}  is a type of SOMs \cite{Fiser:2013} and can find the topology of an input distribution \cite{GarciA-RodriGuez:2012}. The network of the GNG is flexible, and its structure represents the data structure. GNG has been widely applied to topology learning, such as the extraction of the two-dimensional outline of an image \cite{Angelopoulou:2011,Angelopoulou:2018,Fujita:2013}, the reconstruction of 3D models \cite{Holdstein:2008}, landmark extraction \cite{Fatemizadeh:2003}, object tracking \cite{FrezzaBuet:2008}, anomaly detection \cite{Sun:2017}, and cluster analysis \cite{Canales:2007,Costa:2007,Fujita:2021b}.

The kernel method is often used for nonlinear separations. The most famous application of the kernel method is the support vector machine \cite{Cortes:1995,Lee:2010}. Many researchers have used the kernel method to improve the performance of various methods. The kernel $k$-means \cite{Girolami:2002} partitions the data points into a higher dimensional feature space and can partition a dataset non-linearly. Kohonen's SOM has also been kernelized \cite{Aiolli:2007,Andras:2002,Lau:2006}. The kernel Kohonen's SOM shows better performance. However, the kernel GNG is not yet proposed.

% section related_works (end)

\section{Kernel growing neural gas} % (fold)
\label{sec:kernel_gng}

The kernel growing neural gas (kernel GNG) is a modified version of the GNG that uses a kernel function. The kernel GNG projects the dataset into a higher dimensional feature space using a non-linear function and converts it into a network. The kernel trick allows the kernel GNG to learn the topology of the input without the need for direct projection of the data into feature space.

The kernel GNG consists of a set of units connected by a set of unweighted and undirected edges. Each unit $i$ has a weight $\bm{w}_i \in \mathbb{R}^d$ corresponding to a reference vector in the input space and a cumulative error $E_i$. Given a data point from the dataset $X=\{\bm x_1, ..., \bm x_n, ..., \bm x_N\}$, where $\bm x_n \in \mathbb{R}^d$ at each iteration, the kernel GNG updates the unit and the network.

Consider a data point $\bm{x}_n$, a unit weight $\bm{w}_i$, and a nonlinear mapping function $\bm \phi(\cdot)$ that maps $\bm{x}_n$ and $\bm{w}_i$ to $\bm \phi(\bm{x}_n)$ and $\bm \phi(\bm{w}_i)$ in feature space. The dot product of the two points, $\bm \phi(\bm{x}_n)$ and $\bm \phi(\bm{w}_i)$, is denoted as $K(\bm{x}_n, \bm{w}_i) = \bm \phi(\bm{x}_n)^\mathrm{T}\bm \phi(\bm{w}_i)$, where $K(\cdot, \cdot)$ is the kernel function.

The winning unit $s_1$ of the kernel GNG is the one closest to an input data point $\bm x_n$ in the feature space. The criterion to identify $s_1$ is the squared distance between $\bm{x}_n$ and the weight $\bm{w}_{i}$ of unit $i$ in the feature space. The squared distance $D^2(\bm{x}_n, \bm{w}_i)$ in the feature space is defined as follows:
\begin{eqnarray}
    \label{eq:dist}
    D^2(\bm{x}_n, \bm{w}_i) &=& \|\bm \phi(\bm{x}_n) - \bm \phi(\bm{w}_i)\|^2 = K(\bm{x}_n, \bm{x}_n) - 2 K(\bm{x}_n, \bm{w}_i) + K(\bm{w}_i, \bm{w}_i).
\end{eqnarray}
The kernel GNG identifies a winning unit by minimizing the squared distance between the mapped point and the mapped weight using the above equation. The winning unit $s_1$ with respect to an input $\bm{x}_n$ is thus obtained by
\begin{equation}
    s_1 = \arg \min_i (D^2(\bm{x}_n, \bm{w}_i)).
\end{equation}
After that, the weight of the winning unit $\bm{w}_{s_1}$ is updated according to the following rule:
\begin{equation}
    \bm{w}_{s_1}(t + 1) = \bm{w}_{s_1}(t) - \varepsilon_{s_1} \frac{1}{2}\frac{\partial}{\partial \bm{w}_{s_1}} D^2(\bm{x}_n, \bm{w}_{s_1}),
\end{equation}
where $t$ is the iteration index and $\varepsilon_{s_1}$ is the learning rate of the winning unit $s_1$. This equation is based on gradient descent to minimize the squared distance $D^2(\bm{x}_n, \bm{w}_{s_1})$. Therefore, the update equation for the weight $\bm{w}_{s_1}$ in the kernel GNG is as follows
\begin{equation}
    \bm{w}_{s_1}(t + 1) = \bm{w}_{s_1}(t) - \varepsilon_{s_1} \frac{1}{2}\left(\frac{\partial}{\partial \bm{w}_{s_1}}  K(\bm{w}_{s_1}, \bm{w}_{s_1}) - 2 \frac{\partial}{\partial \bm{w}_{s_1}} K(\bm{w}_{s_1}, \bm{x}_n)\right).
\end{equation}
This equation is consistent with that of the kernel SOM update rule proposed by \cite{Andras:2002}. This method eliminates the need to maintain the transformed weights and the transformed data points, allowing direct updating of the weights in the input space without updating the high-dimensional weights in the feature space.

Five kernel functions are used in this study, including Gaussian, Laplacian, Cauchy, inverse multiquadric (IMQ), and log kernels.  Table \ref{tab:kernels} shows the kernelized $D^2(\bm x, \bm w)$ and $\frac{\partial}{\partial \bm{w}} D^2(\bm x, \bm w)$.

A more detailed description of the derivation of the update equation for Gaussian kernels is given in the Appendix. The code for the kernel GNG is openly available on GitHub (\url{https://github.com/KazuhisaFujita/KernelGNG}).

\begin{table}[hb]
\caption{$K(\bm{x}, \bm{w})$, $D^2(\bm{x}, \bm{w})$, and differentiations of $D^2(\bm{x}, \bm{w})$ \label{tab:kernels}}

\begin{tabular}{|c|c|c|c|}
\hline
kernel     &$K(\bm{x}, \bm{w})$
           &$D^2(\bm{x}, \bm{w})$
           &$\frac{\partial}{\partial \bm{w}} D^2(\bm x, \bm w)$\\ \hline
Gaussian   &$\exp(-\frac{\| \bm{x} - \bm{w} \|^2}{2\gamma^2})$
           &$2 ( 1 - \exp (-\frac{\| \bm{x} - \bm{w} \|^2}{2\gamma^2} ) )$
           &$-2 \frac{\bm x - \bm w}{\gamma^2} \exp(-\frac{\| \bm{x} - \bm{w} \|^2}{2\gamma^2} )$\\ \hline
Laplacian  &$\exp(-\frac{\| \bm{x} - \bm{w} \|}{\gamma})$
           &$2(1-\exp(-\frac{\|\bm x - \bm w\|}{\gamma}))$
           &$-\frac{2}{\gamma} \frac{\bm x - \bm w}{\|\bm x - \bm w\|} \exp (-\frac{\| \bm{x} - \bm{w} \|}{\gamma} ) $\\ \hline
Cauchy     &$\frac{1}{1+\|\bm{x} - \bm{w}\|^2/\gamma^2}$
           &$2 (1 - \frac{1}{1+\|\bm{x} - \bm{w}\|^2/\gamma^2})$
           &$- \frac{4}{\gamma^2}\frac{\bm x - \bm w}{(1+\|\bm{x} - \bm{w}\|^2/\gamma^2)^2}$\\ \hline
IMQ        &$\frac{1}{\sqrt{\|\bm{x} - \bm{w}\|^2 + \gamma^2}}$
           &$ 2(\frac{1}{c} - \frac{1}{\sqrt{\|\bm{x} - \bm{w}\|^2 + \gamma^2}})$
           &$-2\frac{\bm{x} - \bm{w}}{(\|\bm{x} - \bm{w}\|^2 + \gamma^2)^{3/2}}$ \\ \hline
log        &$-\log(\|\bm x - \bm w \|^\gamma + 1)$
           &$2\log(\|\bm x - \bm w\|^\gamma+1)$
           &$-2d(\bm x - \bm w)\frac{\|\bm x - \bm w\|^{\gamma-2}}{\|\bm x - \bm w\|^\gamma+1}$ \\ \hline
    \end{tabular}
\end{table}

\subsection{Algorithm of the kernel GNG} % (fold)
\label{sub:algorithm_of_kernel_gng}

The kernel GNG, based on the same principles as the original GNG algorithm, extracts the network structure from the input data, but uses kernelized equations. The algorithm is formulated as
\begin{enumerate}
    \item Initialize the network with two connected neurons. Their weights are two randomly selected data points.
    \item Randomly select an input data point $\bm{x}_n$ from the dataset.
    \item Identify the winning unit $s_1$, the one closest to $\bm{x}_n$, as defined by
        \begin{eqnarray}
            s_1 &=& \arg \min_i D^2(\bm{x}_n, \bm{w}_i),
        \end{eqnarray}
        where $D^2$ is shown in table \ref{tab:kernels}. At the same time, find the second nearest unit, $s_2$.
    \item Increment the ages of all edges connected to the winning unit $s_1$.
    \item Increase the cumulative error $E_{s_1}(t)$ by the squared distance between the input data point $\bm{x}_i$ and the weight of the winning unit $\bm{w}_{s_1}$:
        \begin{equation}
            E_{s_1}(t + 1) = E_{s_1}(t) + D^2(\bm{x}_n, \bm{w}_{s_1}).
        \end{equation}
    \item Adapt the winning unit $s_1$ and its neighbors $j$ to better reflect the input data point $\bm{x}_n$ by updating their weights:
        \begin{equation}
            \bm{w}_{s_1}(t + 1) = \bm{w}_{s_1}(t) - \varepsilon_{s_1} \frac{1}{2}\frac{\partial}{\partial \bm{w}_{s_1}} D^2(\bm{x}_n, \bm{w}_{s_1}),
        \end{equation}
        \begin{equation}
            \bm{w}_{j}(t + 1) =  \bm{w}_{j}(t) - \varepsilon_{n} \frac{1}{2}\frac{\partial}{\partial \bm{w}_j} D^2(\bm{x}_n, \bm{w}_j),
        \end{equation}
        where $\frac{\partial}{\partial \bm{w}} D^2(\bm{x}, \bm{w})$ is in table \ref{tab:kernels}.
    \item If the units $s_1$ and $s_2$ are connected, reset the age of their connecting edge to zero. Otherwise, create an edge between them.
    \item Discard any edges whose ages exceed the maximum age $a_\mathrm{max}$. If this leaves any units isolated, remove them.
    \item Insert a new unit after every $\lambda$ iteration:
    \begin{itemize}
        \item Identify the unit $q$ with the largest cumulative error $E_q$.
        \item Among the neighbors of $q$, find the node $f$ with the largest error.
        \item Insert a new unit $r$ between $q$ and $f$ as follows:
        \begin{equation}
            \bm{w}_r = (\bm{w}_q + \bm{w}_f)/2.
        \end{equation}
        \item Create edges between neurons $r$ and $q$, and between $r$ and $f$, while removing the edge between $q$ and $f$.
        \item Decrease the cumulative errors of $q$ and $f$ by multiplying them by a constant $\alpha$, and initialize the cumulative error of $r$ to the updated error of $q$.
    \end{itemize}
    \item Multiply all cumulative errors by a constant $\beta$ to reduce them.
    \item Repeat from step 2 until the number of iterations reaches $T$.
\end{enumerate}
In this study we used $N_\mathrm{max} = 100$, $a_\mathrm{max} = 50$, $\lambda = 100$, $\alpha = 0.5$, $\beta = 0.995$, $\varepsilon_{s_1} = 0.2$, and $\varepsilon_n = 0.006$. These parameter settings are based on \cite{Fritzke:1995}.

\section{Evaluation metrics for kernel GNG performance and network topology}

\subsection{Evaluation metrics for kernel GNG}

The effectiveness of the kernel GNG is evaluated using two different metrics. 

The first metric, mean square error (MSE), evaluates the average of the squared distances between each input data point and its nearest unit in the input space. This metric is expressed as
\begin{equation}
    \mathrm{MSE} = \sum_{n = 1}^N \min_i \|\bm x_n - \bm w_i \|^2,
\end{equation}
where $\bm x_n$ is an input data point, $\bm w_i$ is the weight of neuron $i$, and $\|.\|$ is the Euclidean norm. 

The second metric, kernel mean square error (kMSE), extends the MSE by measuring the squared distance between the data points and their corresponding weight vectors in the transformed feature space defined by the kernel function. The kMSE is expressed as
\begin{equation}
    \mathrm{kMSE} = \sum_{n=1}^N \min_i D^2(\bm x_n, \bm w_i).
\end{equation}
In the above equation, $D^2(\bm{x}_n, \bm{w}_i)$ is the distance metric computed in the feature space between the input data point $\bm{x}_n$ and the weight vector $\bm{w}_i$. For more information about the kernel distance metric $D^2(\bm{x}_n, \bm{w}_i)$, see table \ref{tab:kernels}.

\subsection{Network analysis metrics for topology evaluation} % (fold)
\label{sec:features_of_graph_structure}

In the field of complex network research, measures are used to study the structure of networks. In this study, two measures are used to examine the generated networks: the average degree and the average clustering coefficient. 

The average degree, denoted as $k$, quantifies the average number of edges per node. It is calculated using the following formula
\begin{equation}
k = \frac{1}{N} \sum_{i = 1}^N k_i,
\end{equation}
where $k_i$ is the degree (the number of edges) of node $i$.

On the other hand, the average clustering coefficient $C$ indicates how much nodes in the network tend to form connected triangles on average. It is given by
\begin{equation}
C = \frac{1}{N} \sum_{i = 1}^{N} c_i,
\end{equation}
where $c_i$ is the clustering coefficient of node $i$. The clustering coefficient $c_i$ for node $i$ \cite{Saramaki:2007} is given by
\begin{equation}
c_i = \frac{2t_i}{k_i (k_i - 1)},
\end{equation}
where $t_i$ is the number of triangles around $i$ and $k_i$ is the degree of $i$. If $k_i < 2$, $c_i$ is set to zero.

These metrics are derived using NetworkX, a comprehensive Python library tailored for network analysis.

\section{Experimental setting}

For this research, we used several Python libraries, namely NumPy for calculations related to linear algebra, NetworkX for handling network operations and computing coefficients, and scikit-learn for generating synthetic data.

Synthetic and real-world data sets are used to evaluate the characteristics of the network generated by kernel GNG. The synthetic datasets include Square, Blobs, Circles, Moons, Swiss\_roll, and S\_curve. Square dataset is constructed using NumPy's \textit{random.rand} function, which generates two-dimensional data points uniformly distributed between 0 and 1. Blobs dataset is constructed using scikit-learn's \textit{datasets.make\_blobs} function, which uses a Gaussian mixture model of three isotropic Gaussian distributions with default parameters. Circles dataset, created using the \textit{datasets.make\_circles} function with noise and scale parameters of 0.05 and 0.5, respectively, contains two concentric circles of data points. The Moons dataset, a distribution mimicking the shape of crescents, was created using the \textit{datasets.make\_moons} function with a noise parameter of 0.05. Swiss\_roll and S\_curve datasets are generated using \textit{datasets.make\_swiss\_roll} and \textit{datasets.make\_s\_curve}, respectively. Each synthetic dataset contains 1000 data points. In addition to these synthetic datasets, we also use two-dimensional datasets such as Aggregation \cite{Gionis:2007}, Compound \cite{Zahn:1971}, Pathbased \cite{Chang:2008}, Spiral \cite{Chang:2008}, D31 \cite{Veenman:2002}, R15 \cite{Veenman:2002}, Jain \cite{Jain:2005}, Flame \cite{Fu:2007}, and t4.8k \cite{Karypis:1999}. The real-world datasets used are Iris, Wine, Ecoli, Glass, Yeast, Spam, CNAE-9, and Digits from the UCI Machine Learning Repository.

In all experiments performed, the preprocessing includes normalizing each data point $\bm{x}_n = (x_{n1}, ..., x_{nd}, ..., x_{nD})$ in a dataset $X = \{\bm x_1, ..., \bm x_n, ..., \bm x_N\}$ using the following formula:
\begin{equation}
    \bm{x}_n = \bigl(\frac{x_{n1} - \bar{x}_1}{\sigma_1}, ..., \frac{x_{nd} - \bar{x}_d}{\sigma_d}, ..., \frac{x_{nD} - \bar{x}_D}{\sigma_D}\bigr),
\end{equation}
where $\bar{x}_d = \frac{1}{N}\sum_{n=1}^N x_{nd}$, and $\sigma_{d} = \sqrt{\frac{1}{N} \sum_{n=1}^N (x_{nd} - \bar{x}_d)^2}$.

\section{Results} % (fold)
\label{sec:results}

In this section, we present four experimental results that demonstrate the effectiveness of kernel GNGs. First, we provide a 2-dimensional visualization of the networks generated by kernel GNGs, illustrating their structure and connectivity. Second, we present the evolution of MSE and kernel MSE over iterations $t$. Third, we explore the dependence of MSE and network structure on the kernel parameter $\gamma$, revealing the role of this parameter in shaping the network. Finally, we describe the characteristics of the network generated by kernel GNGs.

\subsection{Visualization of networks generated by GNG and Kernel GNGs} % (fold)
\label{sub:synthetic_dataset}

\begin{figure}[tb]
  \begin{center}
    \includegraphics[width=120mm]{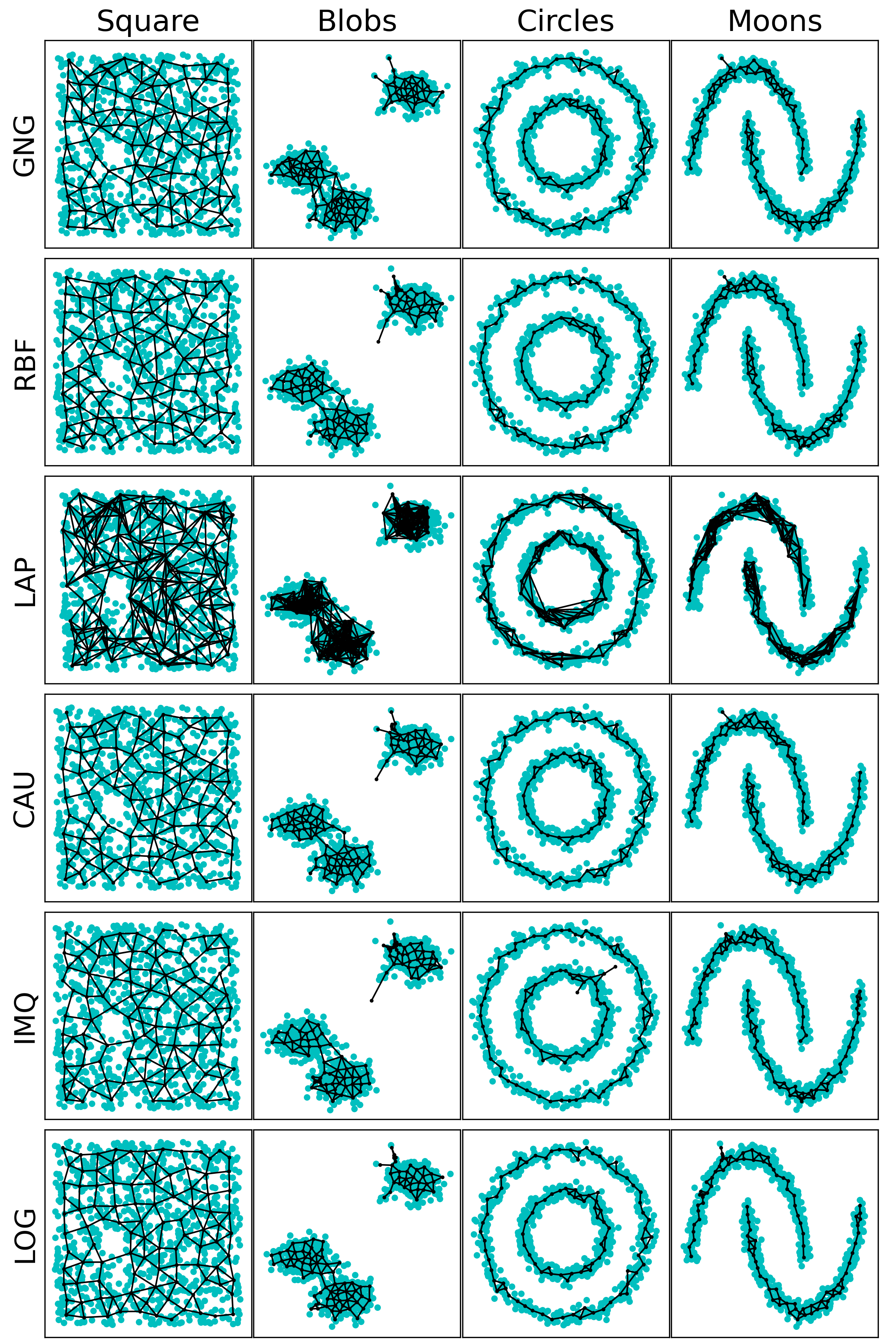}
    \caption{This figure shows networks generated from synthetic data by the GNG and the kernel GNGs with Gaussian, Laplacian, IMQ, Cauchy, and log kernels. Data points are shown as gray dots, while the network units are shown as black dots, with their positions indicating the reference vectors. Black lines mark the edges of the networks.}
    \label{fig:network}
  \end{center}
\end{figure}

Figure \ref{fig:network} shows the networks generated from synthetic datasets by the GNG and kernel GNGs. The kernels used for the kernel GNGs include the Gaussian kernel with $\gamma = 1.8$, the Laplacian kernel with $\gamma = 1.8$, the IMQ kernel with $\gamma = 1.8$, the Cauchy kernel with $\gamma = 1.8$, and the log kernel with $\gamma = 3$. All networks are derived with $\mathrm{END} = 2\times 10^4$, and the random seed is set to 1.

In all cases, the networks generated by kernel GNGs accurately reflect the input topology. However, the Laplacian kernel produces a significantly more complex network structure, but it spreads the units over the input topology. This result shows the ability of kernel GNGs to effectively extract the topology of the dataset.

\subsection{Convergence of MSE and kMSE for Kernel GNGs} % (fold)

\begin{figure*}[tb]
    \centering
      \includegraphics[width=120mm]{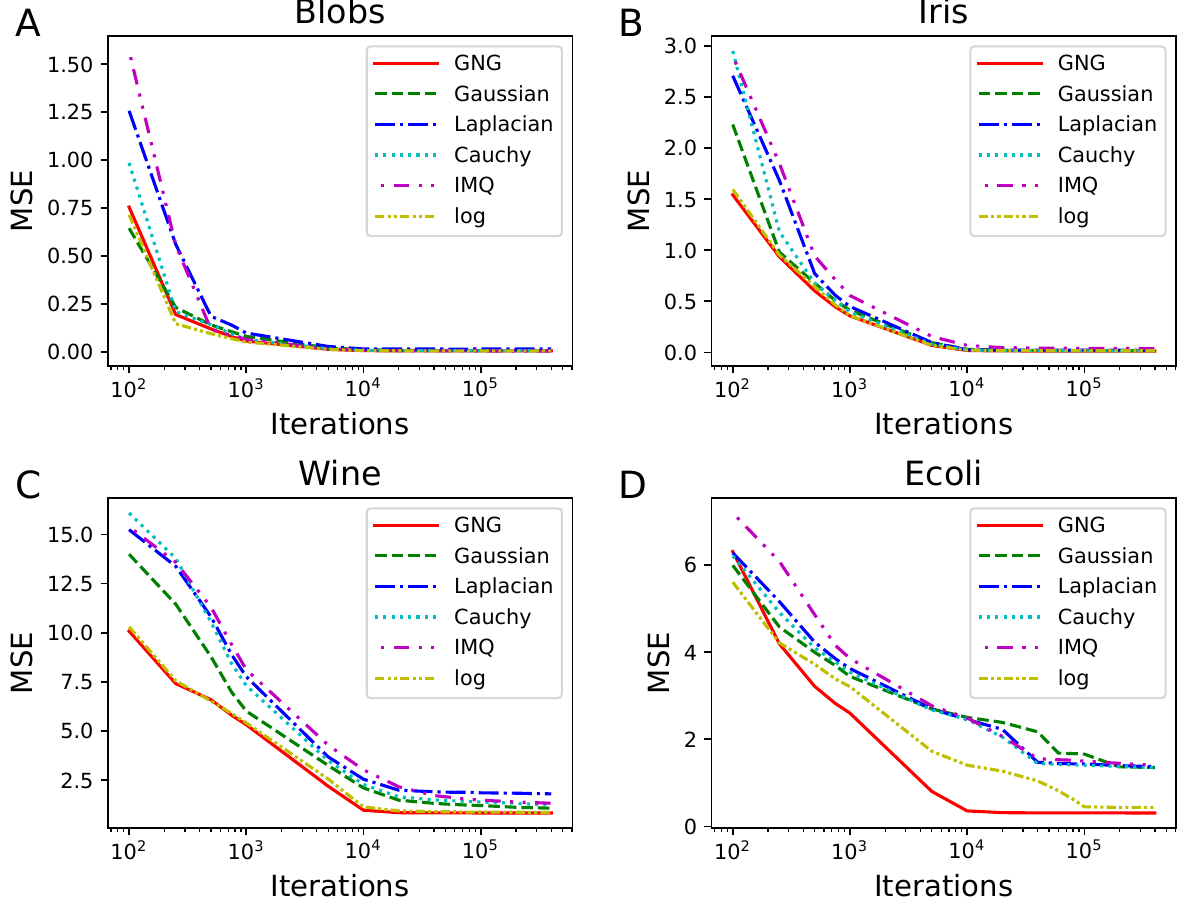}
      \caption{
        Evolution of the MSE over iterations. Subfigures (A) to (D) show the MSE for different data sets: (A) Blobs, (B) Iris, (C) Wine, and (D) Ecoli, respectively. Each data point represents the average kMSE obtained from 10 independent runs, all initialized with random values.
          }
      \label{fig:mse}
  \end{figure*}

Figure \ref{fig:mse} shows the convergence patterns of both the MSE of the kernel GNG and the GNG over iterations. For Blobs and Iris, the MSEs corresponding to all kernel GNGs begin to converge at about $10^4$ iterations. In contrast, for Wine, the MSE of the kernel GNG using the Laplacian and the log starts to converge at about $2 \times 10^4$ iterations, while the others continuously and slowly decrease even after $2 \times 10^4$ iterations. For Wine, the MSEs of the kernel GNGs are larger than that of the GNG, except when the log kernel is used. For Ecoli, the MSEs associated with the kernel GNGs with the Laplacian, Cauchy, and IMQ kernels converge at $4 \times 10^4$ iterations. The kernel GNGs with Gaussian and logarithmic kernels converge at $2 \times 10^5$ and $10^5$ iterations, respectively. For Ecoli, the convergence values of the kernel GNGs with all kernels are larger than those of the GNG. 

\begin{figure*}[tb]
    \centering
      \includegraphics[width=120mm]{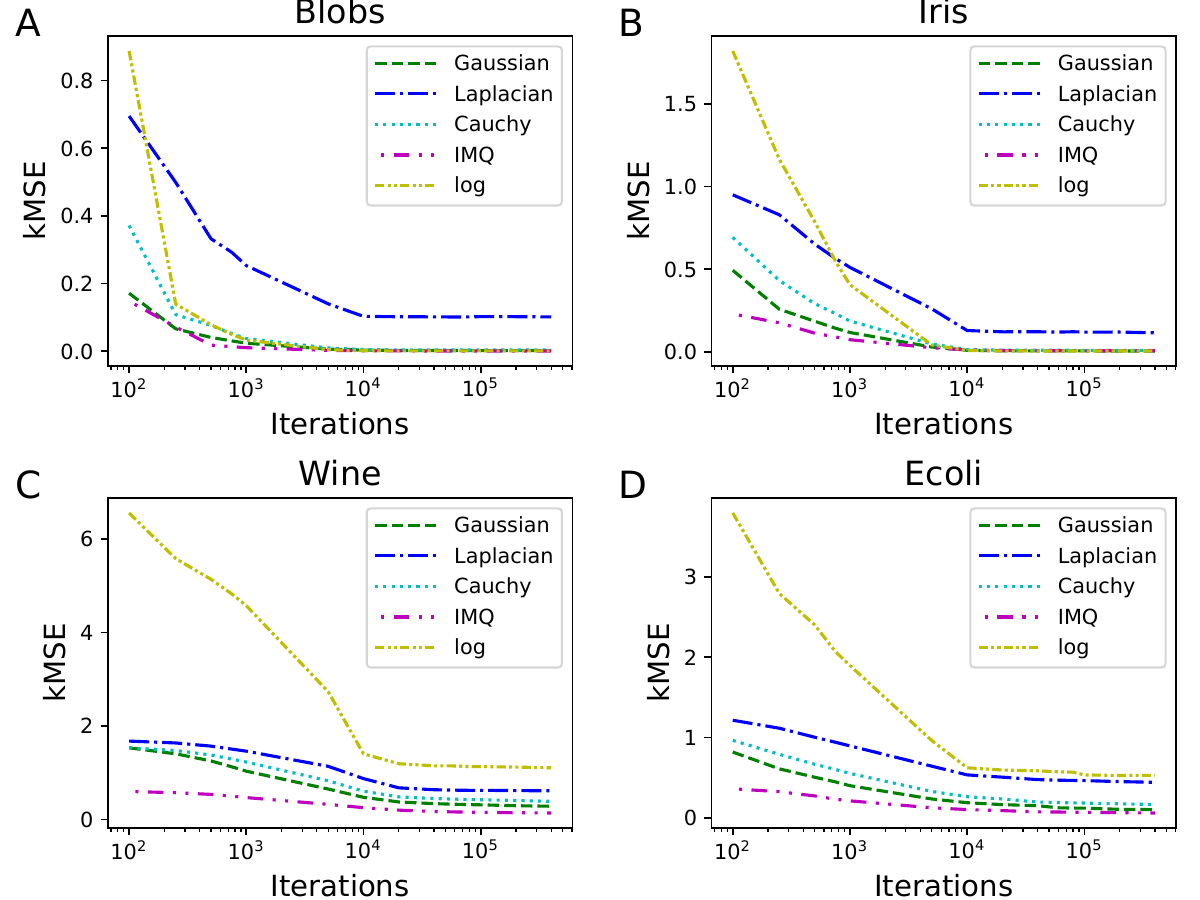}
      \caption{
        Evolution of the kernel mean square error (kMSE) over iterations. Subfigures (A) to (D) show the kMSE for different data sets: (A) Blobs, (B) Iris, (C) Wine, and (D) Ecoli, respectively. The values shown are the averages derived from 10 independent runs, each initialized with random values.
        }
      \label{fig:kmse}
  \end{figure*}

Figure \ref{fig:kmse} shows the convergence behavior of the kMSE over iterations for the kernel GNGs. For Blobs and Iris, the kMSE for all kernel GNGs starts to converge at about $10^4$ iterations. In contrast, for Wine, the kMSE approaches a low value at about $2 \times 10^4$ iterations. For Ecoli, while the kMSE reaches low values at $10^4$ iterations, it exhibits a slow and continuous decline after this iteration. These observations suggest that the kernel GNGs reach appropriately low MSE and kMSE values around $2 \times 10^4$ iterations.

\subsection{Influence of kernel parameters on network characteristics in kernel GNGs} % (fold)

\begin{figure*}[tb]
    \centering
      \includegraphics[width=140mm]{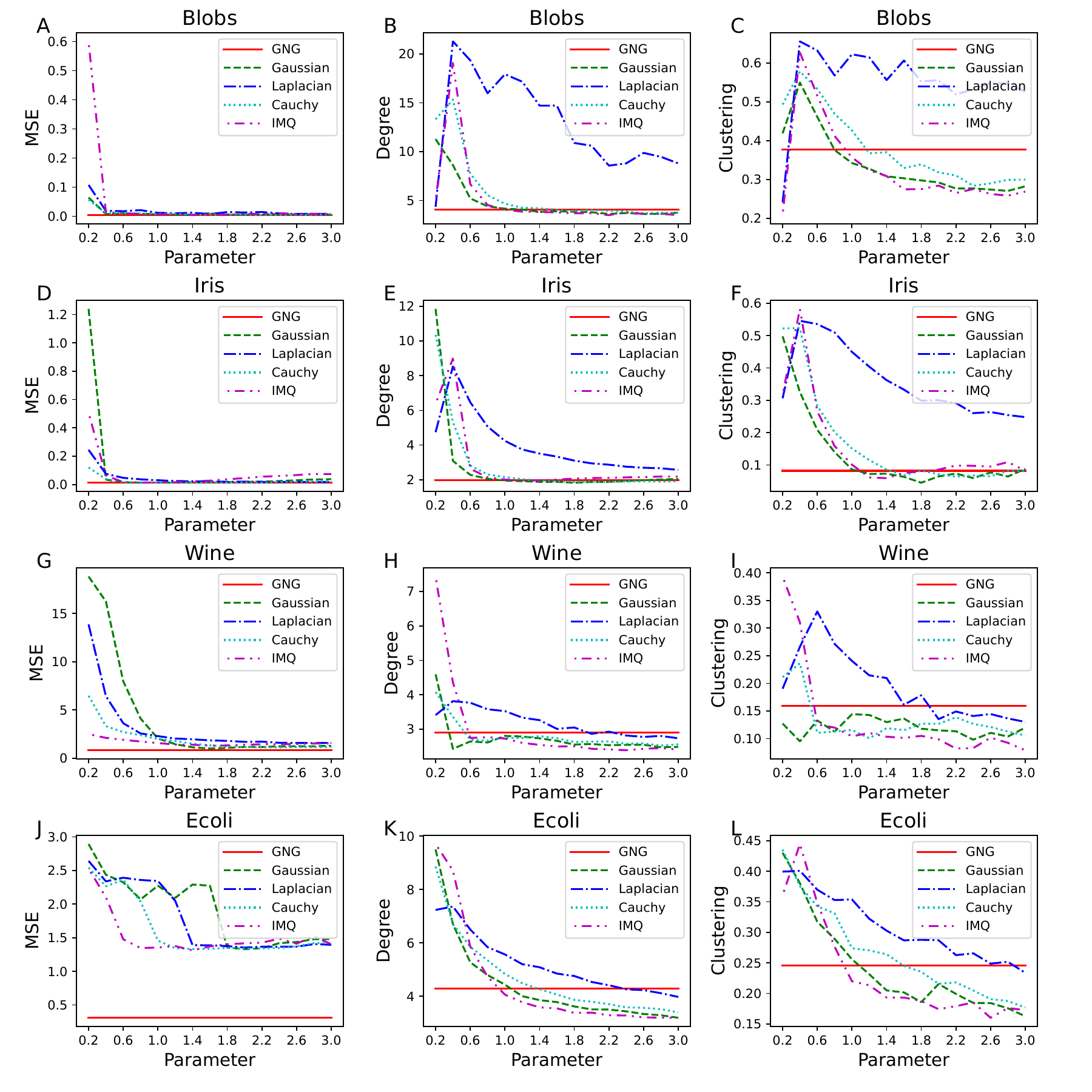}
      \caption{Dependence of various network metrics on kernel parameters for networks generated by kernel GNG with Gaussian, Laplacian, Cauchy, and IMQ kernels for Blobs, Iris, Wine, and Ecoli datasets. Subfigures (A), (D), (G), and (J) show the dependence of the MSE on the kernel parameter. Subfigures (B), (E), (H), and (K) show the average degree, while subfigures (C), (F), (I), and (L) show the average clustering coefficient. Each value shown represents an average derived from 10 independent runs, each initialized with random values.}
      \label{fig:param}
\end{figure*}

Figure \ref{fig:param} shows the dependence of the kernel parameters on the MSE, the average degree, and the average clustering coefficient of the networks generated by the kernel GNG with Gaussian, Laplacian, Cauchy, and IMQ kernels. The MSE, average degree, and average clustering coefficient are computed from the network at $4 \times 10^5$ iterations. Interestingly, as the kernel parameters increase, the MSE, the average degree, and the average clustering coefficient decrease. The kernel GNG with the Laplacian kernel tends to generate a network characterized by a higher degree and a higher clustering coefficient than the others.

\begin{figure*}[tb]
    \centering
      \includegraphics[width=140mm]{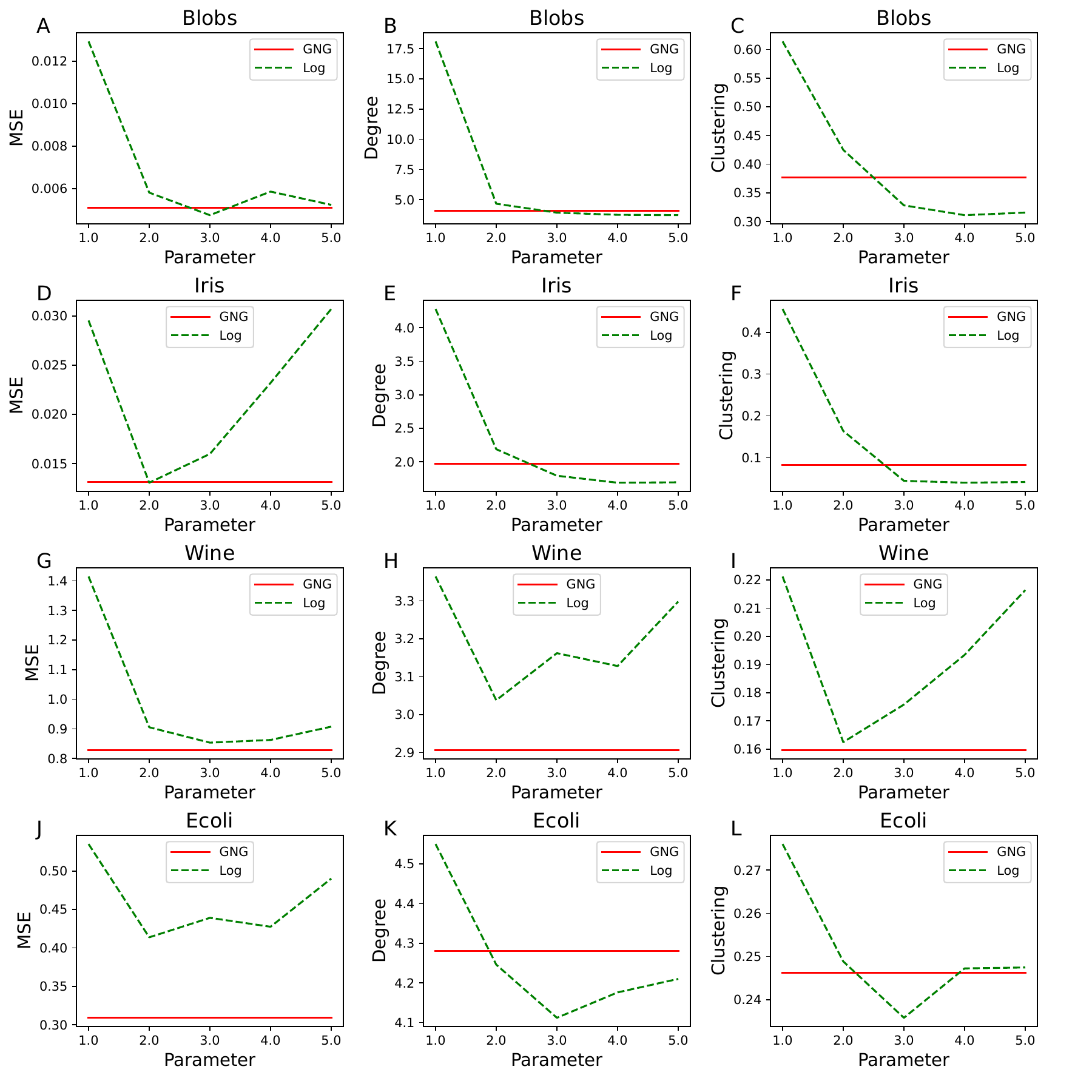}
      \caption{Figure \ref{fig:param_log}: Dependence of various network metrics on kernel parameters in networks generated by kernel GNG with log kernel and GNG for Blobs, Iris, Wine, and Ecoli datasets. Subfigures (A), (D), (G), and (J) show the influence of kernel parameters on the MSE. Subfigures (B), (E), (H), and (K) show the average degree, while subfigures (C), (F), (I), and (L) show the average clustering coefficient. Each displayed value represents the average derived from 10 independent runs, each initialized with random values.}
      \label{fig:param_log}
\end{figure*}

Figure \ref{fig:param_log} shows the effect of the kernel parameters on the MSE, the average degree, and the average clustering coefficient of the networks generated by the kernel GNG with log kernel, with the computations ending at the $4 \times 10^5$ iterations. A noteworthy observation is the absence of any discernible dependence of the network features on the kernel parameter of the kernel GNG with log kernel.

\subsection{Comparison of network characteristics in kernel GNGs}
\label{subsec:characteristics}

Table \ref{tab:edges} gives a comprehensive overview of the average degree $N_d$ of the networks generated by the GNG and the kernel GNGs with the Gaussian, Laplacian, Cauchy, IMQ, and log kernels. The Gaussian, Laplacian, Cauchy, and IMQ kernels use a $\gamma$ parameter value of 1.8. In contrast, the log kernel uses a $\gamma$ parameter value of 3. Each row represents a dataset, and the corresponding data dimension is documented in the second column, denoted by $D$. For networks generated by the kernel GNG using the Gaussian, Cauchy, and IMQ kernels, $N_d$ is less than or equal to that of GNG. While the Laplacian kernel often produces a larger $N_d$ than the GNG, especially for two-dimensional datasets, it produces a smaller $N_d$ for datasets such as CNAE and Digits. In many cases, the log kernel's $N_d$ is equal to or smaller than the GNG's, although it is larger than the GNG's values for datasets such as Wine, Spam, Glass, Yeast, and Digits.

In parallel, table \ref{tab:clustering} shows the average clustering coefficient $C$ of the networks generated by the GNG and the kernel GNGs. The values of $C$ for the kernel GNGs using Gaussian, Cauchy, and IMQ kernels are often less than or equal to those of the GNG. The Laplacian kernel often produces a $C$ larger than the GNG's for lower dimensional data. In addition, the log kernel's $C$ tends to be higher than the other kernels for data points with larger dimensions.

In summary, Gaussian and Cauchy kernels typically produce networks of equal or reduced complexity compared to the GNG. The IMQ kernel tends to produce simpler networks, the Laplacian kernel produces more complex networks, and the log kernel produces more complex networks, especially for high-dimensional data.

\begin{table}[hb]
\begin{center}
\caption{The average degree of a network, $N_d$, generated by GNG and kernel GNGs \label{tab:edges}}
\begin{tabular}{|c|c||c|c|c|c|c|c|}
\hline
dataset     &$D$ &  GNG          & Gaussian      &  Laplacian    & Cauchy       & IMQ          & Log\\ \hline
Square      & 2   & 4.24         & 3.97          & \textbf{6.58} & 4.09         & \textit{3.92}& 4.18\\ \hline
Blobs       & 2   & 4.09         & 3.85          & \textbf{10.89}& 3.91         & \textit{3.70}& 3.94\\ \hline
Circles     & 2   & 2.78         & 2.63          & \textbf{4.89} & 2.70         & \textit{2.59}& 2.61\\ \hline
Moons       & 2   & 3.06         & 2.83          & \textbf{6.76} & 3.02         & \textit{2.79}& 2.93\\ \hline
Swiss\_roll & 3   & 4.14         & 3.90          & \textbf{4.84} & 4.01         & \textit{3.80}& 4.25\\ \hline
S\_curve    & 3   & 4.19         & 3.98          & \textbf{4.92} & 4.07         & \textit{3.82}& 4.16\\ \hline
Aggregation & 2   & 3.99         & 3.73          & \textbf{7.29} & 3.87         & \textit{3.67}& 3.83\\ \hline
Compound    & 2   & 3.18         & 2.77          & \textbf{6.59} & 3.06         & \textit{2.51}& 2.82\\ \hline
t4.8k       & 2   & 4.09         & 4.05          & \textbf{6.55} & \textit{3.98}& 4.08         & 4.06\\ \hline
Iris        & 4   & 1.97         & 1.83          & \textbf{3.11} & 1.88         & 2.07         & \textit{1.79}\\ \hline
Wine        & 13  & 2.91         & 2.55          & 3.05          & 2.63         & \textit{2.51}& \textbf{3.16}\\ \hline
Spam        & 57  & 11.52        & \textit{10.46}& 11.76         & 11.94        & 11.86        & \textbf{12.02}\\ \hline
CNAE        & 857 & \textbf{8.36}& 2.13          & 4.70          & \textit{2.04}& 2.09         & 7.24\\ \hline
Ecoli       & 7   & 4.28         & 3.62          & \textbf{4.76} & 3.86         & \textit{3.37}& 4.11\\ \hline
Glass       & 9   & 2.49         & \textit{2.19} & \textbf{2.94} & 2.36         & 2.33         & 2.67\\ \hline
Yeast       & 8   & 9.03         & 8.95          & 9.33          & 9.19         & \textit{8.63}& \textbf{11.20}\\ \hline
Digits      & 64  & 5.07         & \textit{4.24} & 4.61          & 5.31         & 5.05         & \textbf{6.45}\\ \hline
\end{tabular}
\begin{tablenotes}
    $D$ indicates the dimension of a data point. The Gaussian, Cauchy, and IMQ kernels are used with a $\gamma$ parameter set to 1.8, while the logarithmic kernel uses a $\gamma$ parameter set to 3. The purities are the mean of 10 runs with random initial values. The largest and the smallest values are bold and italic, respectively.
\end{tablenotes}
\end{center}
\end{table}

\begin{table}[hb]
\begin{center}
    \caption{Average clustering coefficient of a network generated by GNG and kernel GNGs \label{tab:clustering}}
  
\begin{tabular}{|c|c||c|c|c|c|c|c|}
\hline
dataset     &$D$ &          GNG & Gaussian     & Laplacian    & Cauchy       & IMQ          & log\\ \hline
Square      & 2  &         0.32 & 0.28         & \textbf{0.47}& 0.31         & \textit{0.26}& 0.32\\ \hline
Blobs       & 2  &         0.38 & 0.30         & \textbf{0.55}& 0.34         & \textit{0.28}& 0.33\\ \hline
Circles     & 2  &         0.34 & 0.29         & \textbf{0.59}& 0.31         & \textit{0.26}& 0.27\\ \hline
Moons       & 2  &         0.37 & 0.28         & \textbf{0.65}& 0.34         & \textit{0.27}& 0.31\\ \hline
Swiss\_roll & 3  &         0.33 & 0.27         & \textbf{0.41}& 0.30         & \textit{0.26}& 0.34\\ \hline
S\_curve    & 3  &         0.34 & 0.30         & \textbf{0.42}& 0.31         & \textit{0.28}& 0.32\\ \hline
Aggregation & 2  &         0.46 & \textit{0.37}& \textbf{0.62}& 0.41         & 0.38         & 0.42\\ \hline
Compound    & 2  &         0.28 & 0.22         & \textbf{0.55}& 0.27         & \textit{0.18}& 0.23\\ \hline
t4.8k       & 2  &         0.33 & 0.30         & \textbf{0.50}& \textit{0.30}& 0.31         & 0.32\\ \hline
Iris        & 4  &         0.08 & 0.05         & \textbf{0.30}& 0.08         & 0.08         & \textit{0.04}\\ \hline
Wine        & 13 &         0.16 & 0.12         & \textbf{0.18}& 0.13         & \textit{0.11}& 0.18\\ \hline
Spam        & 57 & \textbf{0.28}& \textit{0.25}& 0.27         & 0.27         & 0.26         & \textbf{0.28}\\ \hline
CNAE        & 857&         0.26 & 0.04         & 0.21         & \textit{0.02}& 0.03         & \textbf{0.38}\\ \hline
Ecoli       & 7  &         0.25 & \textit{0.19}& \textbf{0.29}& 0.24         & 0.19         & 0.24\\ \hline
Glass       & 9  &         0.14 & \textit{0.10}& \textbf{0.30}& 0.13         & 0.12         & 0.14\\ \hline
Yeast       & 8  &         0.37 & 0.34         & 0.34         & 0.34         & \textit{0.32}& \textbf{0.38}\\ \hline
Digits      & 64 &         0.30 & 0.25         & \textit{0.23}& 0.29         & 0.27         & \textbf{0.39}\\ \hline
\end{tabular}
\begin{tablenotes}
    $D$ indicates the dimension of a data point. The purities are the mean of 10 runs with random initial values. The largest and the smallest values are bold and italic, respectively.
\end{tablenotes}
\end{center}
\end{table}

\section{Conclusion and Discussion}
\label{sec:conclusion}

This paper describes the kernel growing neural gas (GNG) and investigates the characteristics of the networks it generates. Several kernel functions are tested, including Gaussian, Laplacian, IMQ, Cauchy, and log kernels. The results show that the reference vectors produced by the kernel GNG match the input dataset, which is confirmed by the sufficiently small mean square error (MSE) values. However, the topology of the network generated by kernel GNG depends on the kernel function parameter $\gamma$. The average degree and the average clustering coefficient decrease as $\gamma$ increases for Gaussian, Laplacian, Cauchy, and IMQ kernels. 

The choice between kernel GNG and GNG is complex, mainly because the only discernible difference from our results is in the metrics of the network topology. However, if we avoid more edges and a higher clustering coefficient (or more triangles), kernel GNG with a larger value of $\gamma$ will be more appropriate. Such a feature will be particularly valuable for 3D graphics applications, where the kernel GNG may be able to simplify the mesh structure of polygons for a more efficient rendering.

\appendix
\section*{Appendix}

\section{Derivation of update rules for kernel GNG with Gaussian kernel}

In many machine learning applications, Gaussian kernel is the most used kernel. It can be expressed as
\begin{equation}
K(\bm{x}_n, \bm{w}_i) = \bm \phi(\bm{x}_n)^\mathrm{T}\bm \phi(\bm{w}_i) = \exp\left(-\frac{\| \bm{x}_n - \bm{w}_i \|^2}{2\gamma^2}\right).
\end{equation}
The squared distance in feature space between the vectors $\bm \phi(\bm{x}_n)$ and $\bm \phi(\bm{w}_i)$ is given by
\begin{align}
D^2(\bm{x}_n, \bm{w}_i)
    &= \|\bm \phi(\bm{x}_n) - \bm \phi(\bm{w}_i) \|^2 \\
    &= \bm \phi(\bm{x}_n)^\mathrm{T} \bm \phi(\bm{x}_n) - 2 \bm \phi(\bm{x}_n)^\mathrm{T} \bm \phi(\bm{w}_i) + \bm \phi(\bm{w}_i)^\mathrm{T}\bm \phi(\bm{w}_i) \\
    &= K(\bm{x}_n, \bm{x}_n) - 2K(\bm{x}_n, \bm{w}_i) + K(\bm{w}_i, \bm{w}_i) \\
    &= 2 \left( 1 - \exp\left(-\frac{\| \bm{x}_n - \bm{w}_i \|^2}{2\gamma^2}\right) \right).
\end{align}
It is important to note that this squared distance is modulated by the parameter $\gamma$, which is fundamental to the structure of the Gaussian kernel. Given our derivations above, the unit that minimizes the squared distance $D^2(\bm{x}_n, \bm{w}_{s_1})$, commonly called the "winning unit," can be determined as
\begin{equation}
s_1 = \arg \min_{i} \left(2 \left( 1 - \exp\left(-\frac{\| \bm{x}_n - \bm{w}_{s_1} \|^2}{2\gamma^2}\right) \right) \right).
\end{equation}

The kernel GNG algorithm uses a gradient descent approach to refine the weight $\bm{w}_{s_1}$. As a result, the update equation becomes
\begin{equation}
\bm{w}_{s_1}(t+1) = \bm{w}_{s_1}(t) - \varepsilon_{s_1}\frac{1}{2}\frac{\partial}{\partial \bm{w}_{s_1}} D^2(\bm{x}_n, \bm{w}_{s_1}).
\end{equation}

Continuing the differentiation of $D^2(\bm{x}_n, \bm{w}_{s_1})$ with respect to $\bm{w}_{s_1}$, we get
\begin{align}
\frac{\partial}{\partial \bm{w}_{s_1}} D^2(\bm x_n, \bm{w}_{s_1})
&= \frac{\partial}{\partial \bm{w}_{s_1}} \left(K(\bm{x}_n, \bm{x}_n) - 2K(\bm{x}_n, \bm{w}_{s_1}) + K(\bm{w}_{s_1}, \bm{w}_{s_1})\right) \\
&= - 2 \frac{\partial}{\partial \bm{w}_{s_1}} K(\bm{x}n, \bm{w}_{s_1}) + \frac{\partial}{\partial \bm{w}_{s_1}} K(\bm{w}_{s_1}, \bm{w}_{s_1}).
\end{align}
Taking this into account, the comprehensive update equation for the kernel GNG is given by
\begin{equation}
\bm{w}_{s_1}(t+1) = \bm{w}_{s_1}(t) - \varepsilon_{s_1}\frac{1}{2}\left(\frac{\partial}{\partial \bm{w}_{s_1}} K(\bm{w}_{s_1}, \bm{w}_{s_1}) - 2 \frac{\partial}{\partial \bm{w}_{s_1}} K(\bm{x}_n, \bm{w}_{s_1})\right).
\end{equation}
When using Gaussian kernel, the differentiation of $K(\bm{x}_n, \bm{w}_i)$ with respect to $\bm{w}_{s_1}$ is:
\begin{align}
\frac{\partial}{\partial \bm{w}_{s_1}} K(\bm{x}_n, \bm{w}_i) &= - \frac{2(\bm{x}_n - \bm{w}_{s_1})}{\gamma^2} \exp\left(- \frac{|\bm{x}_n - \bm{w}_{s_1}|^2}{2\gamma^2}\right).
\end{align}
Thus, in the specific context of the kernel GNG using Gaussian kernel, the update equation for the winning unit $s_1$ becomes
\begin{equation}
\bm{w}_{s_1}(t+1) = \bm{w}_{s_1}(t) + \varepsilon_{s_1} \frac{\bm{x}_i - \bm{w}_{s_1}}{\gamma^2} \exp\left(- \frac{\|\bm{x}i - \bm{w}_{s_1}\|^2}{2\gamma^2}\right).
\end{equation}
To use other kernels, we simply replace the Gaussian kernel part.

%\bibliographystyle{plain}%{naturemag2}%{plain}%{naturemag2}%acm}
%\bibliography{som,complexnetwork,clustering,etc}

\begin{thebibliography}{10}

    \bibitem{Aiolli:2007}
    Fabio Aiolli, Giovanni Da~San Martino, Alessandro Sperduti, and Markus
      Hagenbuchner.
    \newblock "kernelized" self-organizing maps for structured data.
    \newblock In {\em {ESANN} 2007, 15th European Symposium on Artificial Neural
      Networks, Bruges, Belgium, April 25-27, 2007, Proceedings}, pages 19--24,
      2007.
    
    \bibitem{Andras:2002}
    P\'eter Andr\'as.
    \newblock Kernel-kohonen networks.
    \newblock {\em International Journal of Neural Systems}, 12(02):117--135, 2002.
    
    \bibitem{Angelopoulou:2018}
    Anastassia Angelopoulou, Jose Garc{\'i}a-Rodr{\'i}guez, Sergio Orts-Escolano,
      Gaurav Gupta, and Alexandra Psarrou.
    \newblock Fast 2d/3d object representation with growing neural gas.
    \newblock {\em Neural Computing and Applications}, 29:903--919, 2018.
    
    \bibitem{Angelopoulou:2011}
    Anastassia Angelopoulou, Alexandra Psarrou, and Jos{\'e}
      Garc{\'i}a-Rodr{\'i}guez.
    \newblock A growing neural gas algorithm with applications in hand modelling
      and tracking.
    \newblock In Joan Cabestany, Ignacio Rojas, and Gonzalo Joya, editors, {\em
      Advances in Computational Intelligence}, pages 236--243, Berlin, Heidelberg,
      2011. Springer Berlin Heidelberg.
    
    \bibitem{Cabanes:2012}
    Gu\'eNa\"eL Cabanes, Youn\`{e}S Bennani, and Dominique Fresneau.
    \newblock Enriched topological learning for cluster detection and
      visualization.
    \newblock {\em Neural Networks}, 32:186--195, 2012.
    
    \bibitem{Canales:2007}
    Fernando Canales and Max Chac{\'o}n.
    \newblock Modification of the growing neural gas algorithm for cluster
      analysis.
    \newblock In Luis Rueda, Domingo Mery, and Josef Kittler, editors, {\em
      Progress in Pattern Recognition, Image Analysis and Applications}, pages
      684--693, 2007.
    
    \bibitem{Chang:2005}
    Chip-Hong Chang, Pengfei Xu, Rui Xiao, and T.~Srikanthan.
    \newblock New adaptive color quantization method based on self-organizing maps.
    \newblock {\em IEEE Transactions on Neural Networks}, 16:237--249, 2005.
    
    \bibitem{Chang:2008}
    Hong Chang and Dit-Yan Yeung.
    \newblock Robust path-based spectral clustering.
    \newblock {\em Pattern Recognition}, 41(1):191 -- 203, 2008.
    
    \bibitem{Chen:2009a}
    Ning Chen and Hongyi Zhang.
    \newblock Extended kernel self-organizing map clustering algorithm.
    \newblock In {\em 5th International Conference on Natural Computation, ICNC
      2009}, pages 454--458, 2009.
    
    \bibitem{Cortes:1995}
    Corinna Cortes and Vladimir Vapnik.
    \newblock Support-vector networks.
    \newblock In {\em Machine Learning}, pages 273--297, 1995.
    
    \bibitem{Costa:2007}
    J.~A.~F. Costa and R.~S. Oliveira.
    \newblock Cluster analysis using growing neural gas and graph partitioning.
    \newblock In {\em Proceedings of 2007 International Joint Conference on Neural
      Networks}, pages 3051--3056, 2007.
    
    \bibitem{Fatemizadeh:2003}
    E.~Fatemizadeh, C.~Lucas, and H.~Soltanian-Zadeh.
    \newblock Automatic landmark extraction from image data using modified growing
      neural gas network.
    \newblock {\em IEEE Transactions on Information Technology in Biomedicine},
      7(2):77--85, 2003.
    
    \bibitem{Fiser:2013}
    Daniel Fi\~ser, Jan Faigl, and Miroslav Kulich.
    \newblock Growing neural gas efficiently.
    \newblock {\em Neurocomputing}, 104:72--82, 2013.
    
    \bibitem{FrezzaBuet:2008}
    Hervé Frezza-Buet.
    \newblock Following non-stationary distributions by controlling the vector
      quantization accuracy of a growing neural gas network.
    \newblock {\em Neurocomputing}, 71(7–9):1191--1202, 2008.
    
    \bibitem{Fritzke:1995}
    Bernd Fritzke.
    \newblock A growing neural gas network learns topologies.
    \newblock In {\em Proceedings of the 7th International Conference on Neural
      Information Processing Systems}, NIPS’94, page 625–632, Cambridge, MA,
      USA, 1994. MIT Press.
    
    \bibitem{Fu:2007}
    Limin Fu and Enzo Medico.
    \newblock Flame, a novel fuzzy clustering method for the analysis of dna
      microarray data.
    \newblock {\em BMC bioinformatics}, 8:3, 02 2007.
    
    \bibitem{Fujita:2013}
    Kazuhisa Fujita.
    \newblock Extract an essential skeleton of a character as a graph from a
      character image.
    \newblock {\em International Journal of Computer Science Issues}, 10(3):35--39,
      2013.
    
    \bibitem{Fujita:2021b}
    Kazuhisa Fujita.
    \newblock Approximate spectral clustering using both reference vectors and
      topology of the network generated by growing neural gas.
    \newblock {\em PeerJ Comput. Sci.}, 7:e679, 2021.
    
    \bibitem{GarciA-RodriGuez:2012}
    Jos{\'e} Garc\'{\i}a-Rodr\'{\i}Guez, Anastassia Angelopoulou, Juan~Manuel
      Garc\'{\i}a-Chamizo, Alexandra Psarrou, Sergio Orts~Escolano, and Vicente
      Morell~Gim{\'e}Nez.
    \newblock Autonomous growing neural gas for applications with time constraint:
      Optimal parameter estimation.
    \newblock {\em Neural Networks}, 32:196--208, 2012.
    
    \bibitem{Gionis:2007}
    Aristides Gionis, Heikki Mannila, and Panayiotis Tsaparas.
    \newblock Clustering aggregation.
    \newblock {\em ACM Trans. Knowl. Discov. Data}, 1(1):1–30, 2007.
    
    \bibitem{Girolami:2002}
    Mark Girolami.
    \newblock Mercer kernel-based clustering in feature space.
    \newblock {\em IEEE Transactions on Neural Networks}, 13:780--784, 2002.
    
    \bibitem{Heskes:2001}
    Tom Heskes.
    \newblock Self-organizing maps, vector quantization, and mixture modeling.
    \newblock {\em IEEE Transactions on Neural Networks}, 12:12--1299, 2001.
    
    \bibitem{Holdstein:2008}
    Y.~Holdstein and A.~Fischer.
    \newblock Three-dimensional surface reconstruction using meshing growing neural
      gas ({MGNG}).
    \newblock {\em The Visual Computer}, 24(4):295--302, 2008.
    
    \bibitem{Jain:2005}
    Anil~K. Jain and Martin H.~C. Law.
    \newblock Data clustering: A user's dilemma.
    \newblock In Sankar~K. Pal, Sanghamitra Bandyopadhyay, and Sambhunath Biswas,
      editors, {\em Pattern Recognition and Machine Intelligence}, pages 1--10,
      Berlin, Heidelberg, 2005. Springer Berlin Heidelberg.
    
    \bibitem{Karypis:1999}
    George Karypis, {Eui Hong} Han, and Vipin Kumar.
    \newblock Chameleon: Hierarchical clustering using dynamic modeling, 1999.
    
    \bibitem{Kohonen:1982}
    Teuvo Kohonen.
    \newblock Self-organized formation of topologically correct feature maps.
    \newblock {\em Biological Cybernetics}, 43:59--69, 1982.
    
    \bibitem{Lau:2006}
    K.~W. Lau, H.~Yin, and S.~Hubbard.
    \newblock Kernel self-organising maps for classification.
    \newblock {\em Neurocomputing}, 69:2033--2040, 2006.
    
    \bibitem{Lee:2010}
    Ming-Chang Lee and To~Chang.
    \newblock Comparison of support vector machine and back propagation neural
      network in evaluating the enterprise financial distress.
    \newblock {\em International Journal of Artificial Intelligence \&
      Applications}, 1:31--43, 2010.
    
    \bibitem{Rasti:2011}
    J.~Rasti, A.~Monadjemi, and A.~Vafaei.
    \newblock Color reduction using a multi-stage kohonen self-organizing map with
      redundant features.
    \newblock {\em Expert Systems with Applications}, 38(10):13188--13197, 2011.
    
    \bibitem{Rossi:2014}
    Fabrice Rossi.
    \newblock How many dissimilarity/kernel self organizing map variants do we
      need?
    \newblock In Thomas Villmann, Frank-Michael Schleif, Marika Kaden, and Mandy
      Lange, editors, {\em Advances in Self-Organizing Maps and Learning Vector
      Quantization}, pages 3--23, Cham, 2014. Springer International Publishing.
    
    \bibitem{Saramaki:2007}
    Jari Saram\"{a}ki, Mikko Kivel\"{a}, Jukka-Pekka Onnela, Kimmo Kaski, and
      J\'anos Kert\'esz.
    \newblock Generalizations of the clustering coefficient to weighted complex
      networks.
    \newblock {\em Physical Review E}, 75:027105, 2007.
    
    \bibitem{Singh:2000}
    Rahul Singh, Vladimir Cherkassky, and Nikolaos Papanikolopoulos.
    \newblock Self-organizing maps for the skeletonization of sparse shapes.
    \newblock {\em IEEE Transactions on Neural Networks and Learning Systems},
      11:241--248, 2000.
    
    \bibitem{Sun:2017}
    Qianru Sun, Hong Liu, and Tatsuya Harada.
    \newblock Online growing neural gas for anomaly detection in changing
      surveillance scenes.
    \newblock {\em Pattern Recognition}, 64:187--201, 2017.
    
    \bibitem{Veenman:2002}
    C.J. Veenman, M.J.T. Reinders, and E.~Backer.
    \newblock A maximum variance cluster algorithm.
    \newblock {\em IEEE Transactions on Pattern Analysis and Machine Intelligence},
      24(9):1273--1280, 2002.
    
    \bibitem{Vesanto:2000}
    J.~Vesanto and E.~Alhoniemi.
    \newblock Clustering of the self-organizing map.
    \newblock {\em IEEE Transactions on Neural Networks}, 11:586--600, 2000.
    
    \bibitem{Zahn:1971}
    Charles~T. Zahn.
    \newblock Graph-theoretical methods for detecting and describing gestalt
      clusters.
    \newblock {\em IEEE Trans. Computers}, 20(1):68--86, 1971.
    
    \end{thebibliography}

\end{document}